\documentclass{article}

\PassOptionsToPackage{numbers, compress}{natbib}

\usepackage[final]{neurips_2025}

\usepackage[utf8]{inputenc} 
\usepackage[T1]{fontenc}    
\usepackage{hyperref}       
\usepackage{url}            
\usepackage{booktabs}       
\usepackage{amsfonts}       
\usepackage{nicefrac}       
\usepackage{microtype}      
\usepackage{xcolor}         

\usepackage{graphicx}  
\usepackage{booktabs}  
\usepackage{multirow}  
\usepackage{enumitem}
\usepackage{amsmath}

\title{When One Modality Sabotages the Others: \\ A Diagnostic Lens on Multimodal Reasoning}

\author{
\textbf{Chenyu Zhang}$^{1}$\thanks{Equal contribution. Correspondence to \texttt{chenyu\_zhang@alumni.harvard.edu}.} \quad
\textbf{Minsol Kim}$^{2}$\footnotemark[1] \quad
\textbf{Shohreh Ghorbani}$^{2}$ \quad
\textbf{Jingyao Wu}$^{2}$ \\[3pt]
\textbf{Rosalind Picard}$^{2}$ \quad
\textbf{Patricia Maes}$^{2}$ \quad
\textbf{Paul Pu Liang}$^{2}$ \\
$^1$Harvard University \quad
$^2$MIT Media Lab
}

\begin{document}
\maketitle
\begin{abstract}
Despite rapid growth in multimodal large language models (MLLMs), their reasoning traces remain opaque: it is often unclear which modality drives a prediction, how conflicts are resolved, or when one stream dominates. 
In this paper, we introduce \textit{modality sabotage}---a diagnostic failure mode in which a high-confidence unimodal error overrides other evidence and misleads the fused result.
To analyze such dynamics, we propose a lightweight, model-agnostic evaluation layer that treats each modality as an agent, producing candidate labels and a brief self-assessment used for auditing. A simple fusion mechanism aggregates these outputs, exposing \textit{contributors} (modalities supporting correct outcomes) and \textit{saboteurs} (modalities that mislead). Applying our diagnostic layer in a case study on multimodal emotion recognition benchmarks with foundation models revealed systematic reliability profiles, providing insight into whether failures may arise from dataset artifacts or model limitations. More broadly, our framework offers a diagnostic scaffold for multimodal reasoning, supporting principled auditing of fusion dynamics and informing possible interventions.
\end{abstract}

\section{Introduction}
Multimodal large language models (MLLMs) have advanced rapidly in tasks that combine vision, language, and audio, from answering questions \citep{xiao2021next} to processing social signals \citep{liang2024foundations}. Yet in practice, their decisions remain a black box: users cannot tell \emph{which} stream of data the system relied on, \emph{how} conflicting evidence--e.g., when text, audio, and vision suggest different labels--was resolved, if at all, or whether a single sensor dominated the outcome. Prior work has discussed related issues such as \emph{modality collapse}, where vision–language models over-rely on text \citep{goyal2017making}, and \emph{unimodal bias}, where fusion lets one stream dominate across a dataset \citep{cadene1906rubi, park2025assessing}. In contrast, we highlight a distinct diagnostic failure mode we call \emph{modality sabotage}: instance-level cases where a high-confidence unimodal error not only fails locally but actively overrides other evidence and pulls the fused prediction off-target. Unlike collapse or bias, which describe systematic trends, sabotage is a diagnostic lens on individual decisions, making visible which modality misled the model and when. 
Despite strong progress in multimodal fusion \citep{cheng2023semi,cheng2024mips,cheng2022gsrformer,cheng2024umetts,li2023decoupled,richet2024textualized,wang2023incomplete,zadeh2018multimodal,zhang2023learning,zhou2019exploring} and impressive results from MLLMs in vision–language understanding, visual question answering (VQA), and video processing tasks\citep{lei2024large,liu2023visual,zhu2023minigpt,guo2025stimuvar,lin2023video,tu2023implicit}, current systems mostly emphasize cross-modal feature interaction and modality completion, 
leaving how cues map to constructs and how conflicts are resolved largely unexplored.
Decades of psychology and affective computing show that audio and visual cues carry complementary emotional information \citep{banse1996acoustic,wu2021multimodal,schoneveld2021leveraging}, for example facial expressions correlate with pleasant affect \citep{ekman1979facial} while speech acoustics track arousal \citep{bachorowski1999vocal,russell2003facial,wu2024interspeech}. Yet these studies typically isolate unimodal contributions rather than addressing how models should integrate, arbitrate, or dominate across modalities in multimodal settings. 
We address this need with a simple, transparent, model-agnostic framework that treats each modality as an agent, whose outputs constitute a diagnostic layer that records per-modality votes, confidences, and disagreements, enabling systematic analysis of contributions and failure modes before a final fused decision is made. Specifically, we propose a plug-and-play modality-as-agent fusion that queries text (T), audio (A), vision (V), and their joint view (TAV) separately, then aggregates their predictions into a final decision. The design makes attribution explicit at the instance level, surfacing \emph{contributors} (modalities supporting correct answers) and \emph{saboteurs} (modalities that mislead).

Our contributions are threefold: (i) a lightweight framework that yields instance-level attribution without retraining or architectural changes; (ii) a measurable operationalization of modality sabotage for high-confidence but misleading unimodal outputs; and (iii) dataset- and backbone-dependent reliability profiles that clarify whether failures stem from dataset artifacts or model limitations.

\vspace{-5pt}
\section{Methodology}
\label{sec:method}

\vspace{-5pt}
We evaluate the framework across three widely used multimodal emotion recognition benchmarks (MER~\citep{lian2023mer}, MELD~\citep{poria2018meld}, and IEMOCAP~\citep{busso2008iemocap}) and report unimodal and fused performance, top-$k$ coverage, and sabotage diagnostics.

\begin{figure*}[t]
  \centering
  \includegraphics[width=0.8\textwidth]{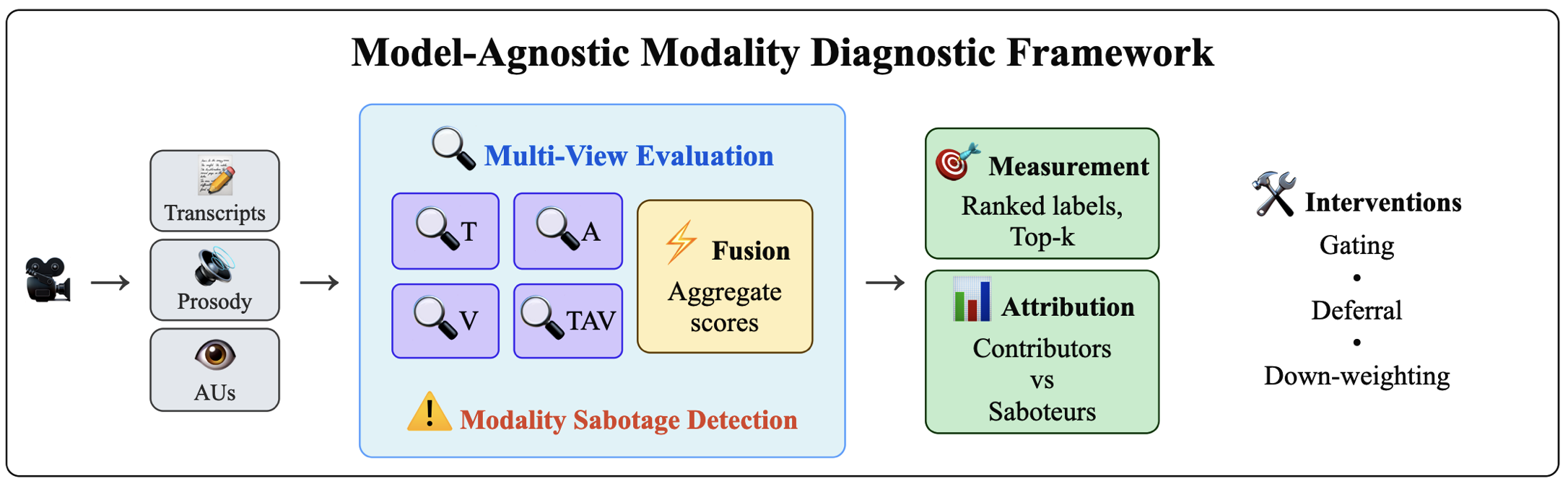}
  \caption{Each modality (T, A, V) and a joint view (TAV) agent outputs classification labels with confidence. A simple fusion aggregates these into a ranked prediction, enabling attribution of \emph{contributors} vs.\ \emph{saboteurs}. 
The callout highlights high-confidence unimodal errors that mislead the fused decision (\emph{modality sabotage}); see Section~\ref{sec:method} for details.}
  \label{fig:overview}
\end{figure*}

\vspace{-6pt}
\paragraph{Inputs per modality.}
For each video segment we derive modality–specific, purely descriptive inputs that avoid direct emotion inference: 
(i) \textbf{Text (T):} Whisper ASR \citep{radford2023robust} transcripts from the audio track serve as the textual input; 
(ii) \textbf{Audio (A):} Each audio utterance instance is analyzed by \textit{Qwen-Audio} \citep{QwenAudio} with a structured prompt to elicit non-lexical descriptors—prosody (pitch/intonation, loudness/intensity, tempo/rhythm), voice quality (breathiness/creak/tension), and articulation—while forbidding use of lexical content or emotion labels; 
(iii) \textbf{Vision (V):} we compute facial AUs with \textit{OpenFace} \citep{baltruvsaitis2016openface}, select an AU-peak frame, and ask a VLM (\textit{GPT-4 Vision} \citep{achiam2023gpt}) to produce an objective caption of observable cues (e.g. facial expressions, posture, gestures, and context) without mental-state attributions. These modality-specific descriptors feed the corresponding modality agents.

\vspace{-10pt}
\paragraph{Agents and outputs.}
We propose a simple, model-agnostic framework (Figure~\ref{fig:overview}) that treats each modality as an \emph{agent} and makes fusion decisions legible. For each sample, T, A, V, and TAV are queried with a structured prompt; each agent returns a sorted set of candidate labels with confidence scores (1--100) together with a \emph{data-quality report} (score 1--100, issues, and a short rationale). 
While the confidence values capture belief over labels, they do not reflect whether the underlying evidence is degraded or incomplete. 
The data-quality report complements confidence by probing whether the LLM can \emph{self-diagnose} potential input issues, such as noisy transcripts, occluded faces, or truncated speech. 
We fuse agents by aggregating their confidences per label and normalizing to obtain a single ranked score vector. 
Let $S_m(y)\in[0,100]$ be the confidence assigned by agent $m\!\in\!\{\mathrm{T},\mathrm{A},\mathrm{V},\mathrm{TAV}\}$ to label $y$ (zero if $y$ is not proposed), and let $q_m\in[0,1]$ denote the agent’s self-reported quality score (rescaled from 1--100). We compute
\[
\tilde{s}(y) \;=\; \sum_{m} w_m \, S_m(y), 
\qquad 
p(y) \;=\; \frac{\tilde{s}(y)}{\sum_{y'} \tilde{s}(y')}\,,
\]
where $w_m=1$ by default and $w_m=q_m$ in a quality-weighted ablation. Across benchmarks, quality weighting did not improve top-1 accuracy (and sometimes reduced it), so we retain the unweighted variant as the main setting and report the weighted variant for completeness. We evaluate using the ranking induced by $p(y)$ and report top-$k$ coverage.

\vspace{-3pt}
\paragraph{Modality sabotage (diagnostic).}
Fusion can fail \emph{silently} when an overconfident stream dominates: a wrong modality can pull the final decision off‑target, and accuracy alone offers no attribution. Let $S_m(y)\!\ge\!0$ denote agent $m$’s evidence for label $y$ (we use self‑reported confidence 1–100; other signals such as probabilities or logits are interchangeable), $p_m(y)=S_m(y)/\sum_{y'}S_m(y')$, $y_m=\arg\max_y p_m(y)$, $c_m=\max_y p_m(y)$, and $\hat{y}=\arg\max_y \tilde{s}(y)$ with $\tilde{s}(y)=\sum_m w_m S_m(y)$.
We distinguish two flavors:
\textbf{Potential sabotage} for $m$ holds when (i) $c_m\!\ge\!\tau$ (high confidence) and (ii) $y_m\!\neq\!y^\ast$ (its own error).
\textbf{Successful sabotage} strengthens this by requiring (iii) $\hat{y}=y_m$ (the fused model follows $m$), with $\tau\!=\!0.70$ unless noted.
However, due to the nature of fusion, successful sabotage does not establish strict causality—multiple agents may jointly support the same wrong label. For this reason, in Section~\ref{sec:results} we focus on \textbf{potential sabotage}, which provides a clearer upper bound on each modality’s tendency toward overconfident errors. Both definitions nonetheless offer actionable diagnostic signals for gating, down-weighting, or deferral.

\vspace{-3pt}
\paragraph{Top-$k$ reasoning.}
Modality sabotage creates a distinctive failure pattern: a single overconfident stream can dominate fusion and force a wrong Top-1 prediction, 
even when other modalities still support the correct label. 
To diagnose whether such errors are recoverable, we evaluate the fused distribution beyond its argmax. 
Specifically, $\mathrm{Acc@}k$ quantifies whether the ground-truth label remains among the top-$k$ hypotheses ranked by the fused scores $p(y)$:
\[
\mathrm{Acc@}k = \frac{1}{N}\sum_{i=1}^{N} \mathbf{1}\!\big[y_i^\ast \in \mathrm{TopK}\big(p_i\big)\big].
\]
Although our test domains (e.g., emotion recognition) have relatively few classes, 
the purpose of Top-$k$ reasoning is not to inflate accuracy through guesswork, 
but to expose \emph{recoverable uncertainty}---cases where the model’s internal ranking still preserves the correct hypothesis despite sabotage.
This diagnostic helps distinguish overconfident collapse (when all alternatives are suppressed) 
from calibrated disagreement (when the true label remains plausible), 
providing actionable signals for gating, abstention, or human review.

\section{Case Study Results}
\label{sec:results}
\vspace{-5pt}
\subsection{Aggregate accuracy and effect of self-reported quality}
\vspace{-2pt}
Table~\ref{tab:agg} compares the single-call \textbf{TAV} baseline, whose Top-1 result is denoted as \textbf{Base T1}, with our \textbf{agentic fusion} (reported as “Fus T1–T5” in the table)  under confidence-only fusion and reports the ablation when additionally weighting by self-reported data quality. Three patterns emerge. \textit{(i) Top}-1 vs. \textit{Top}-$k$. The fusion maintains baseline-level Top-1 on \textbf{MELD} and \textbf{IEMOCAP} and improves markedly on MELD, while Top-$k$ coverage rises steeply across datasets. On \textbf{MER}, Top-1 changes from $0.38$ (baseline) to $0.33$ (fusion, GPT-5-nano), but the correct label appears with high probability in the ranking (Top-5 $=0.97$). On \textbf{MELD}, Top-1 improves by $+0.09$ for GPT-5-nano ($0.27\!\rightarrow\!0.36$) and by $+0.15$ for GPT-4-mini ($0.30\!\rightarrow\!0.45$), with Top-5 $=0.92/0.90$. On \textbf{IEMOCAP}, Top-1 is essentially flat for GPT-5-nano ($0.28\!\rightarrow\!0.29$) and slightly lower for GPT-4-mini ($0.28\!\rightarrow\!0.24$), but Top-5 remains substantially higher than Top-1 (GPT-5-nano: $0.76$, GPT-4-mini: $0.72$).
These results indicate that the fusion retains recoverable uncertainty, preserving the correct hypothesis among its leading options even when the Top-1 prediction is affected by modality conflict.

\vspace{-5pt}
\paragraph{Ablation: confidence $\times$ data quality weighting.}
The $\Delta$ columns quantify the change when scaling each vote by the product of its confidence and self-reported data-quality. Effects are small and often negative: e.g., on \textbf{MELD}/GPT-5-nano, $\Delta$Top-1$=-0.08$ and $\Delta$Top-2$=-0.06$; on \textbf{IEMOCAP}/GPT-5-nano, $\Delta$Top-1$=-0.05$ and $\Delta$Top-3$=-0.07$. Occasional mild gains appear (e.g., \textbf{MER}/GPT-4-mini: $\Delta$Top-4$=+0.02$, $\Delta$Top-5$=+0.03$). 
These findings indicate that self-reported data quality signals capture aspects of model self-perception but are only weakly aligned with correctness. Rather than a weighting mechanism, we view them as a complementary diagnostic signal that may inform future calibration or self-evaluation research.

\begin{table*}[t] 
\centering
\caption{\textbf{Top-$k$ coverage and diagnostic effect of quality weighting.} 
The fusion maintains baseline-level Top-1 accuracy (“Fus T1” vs.\ “Base T1”) while substantially improving Top-$k$ coverage (“Fus T2–T5”). 
The $\Delta$ block reports the change when switching from \emph{confidence-only weighting} to \emph{confidence $\times$ data-quality weighting}.
Comparisons across datasets and backbones (GPT-5-nano vs.\ GPT-4o-mini) highlight systematic differences in modality reliability and pipeline robustness.}
\label{tab:agg}
\resizebox{\linewidth}{!}{%
\begin{tabular}{lcccccc|ccccc}
\toprule
\multirow{2}{*}{Dataset / Model} & \multicolumn{6}{c}{Accuracy} 
& \multicolumn{5}{c}{$\Delta$ (confidence+quality vs. confidence-only)} \\
\cmidrule(lr){2-7}\cmidrule(lr){8-12}
 & Base T1 & Fus T1 & Fus T2 & Fus T3 & Fus T4 & Fus T5 
 & $\Delta$T1 & $\Delta$T2 & $\Delta$T3 & $\Delta$T4 & $\Delta$T5 \\
\midrule
\emph{MER} / GPT-5-nano   & 0.38 & 0.33 & 0.62 & 0.85 & 0.92 & 0.97 & +0.00 & +0.01 & +0.00 & -0.02 & +0.01 \\
\emph{MER} / GPT-4o-mini  & 0.35 & 0.23 & 0.52 & 0.75 & 0.83 & 0.85 & -0.03 & +0.00 & +0.00 & +0.02 & +0.03 \\
\midrule
\emph{MELD} / GPT-5-nano  & 0.27 & 0.36 & 0.58 & 0.73 & 0.86 & 0.92 & -0.08 & -0.06 & -0.03 & -0.03 & -0.04 \\
\emph{MELD} / GPT-4o-mini & 0.30 & 0.45 & 0.64 & 0.76 & 0.85 & 0.90 & -0.02 & +0.01 & -0.02 & -0.02 & -0.02 \\
\midrule
\emph{IEMOCAP} / GPT-5-nano  & 0.28 & 0.29 & 0.47 & 0.62 & 0.73 & 0.76 & -0.05 & -0.07 & -0.07 & -0.02 & +0.03 \\
\emph{IEMOCAP} / GPT-4o-mini & 0.28 & 0.24 & 0.43 & 0.60 & 0.70 & 0.72 & +0.01 & +0.03 & +0.00 & -0.02 & +0.00 \\
\bottomrule
\end{tabular}}
\end{table*}

\vspace{-5pt}
\subsection{Modality behavior and sabotage analysis}
\label{sec:sabotage}
\begin{figure*}[t]
  \centering
  \includegraphics[width=\textwidth]{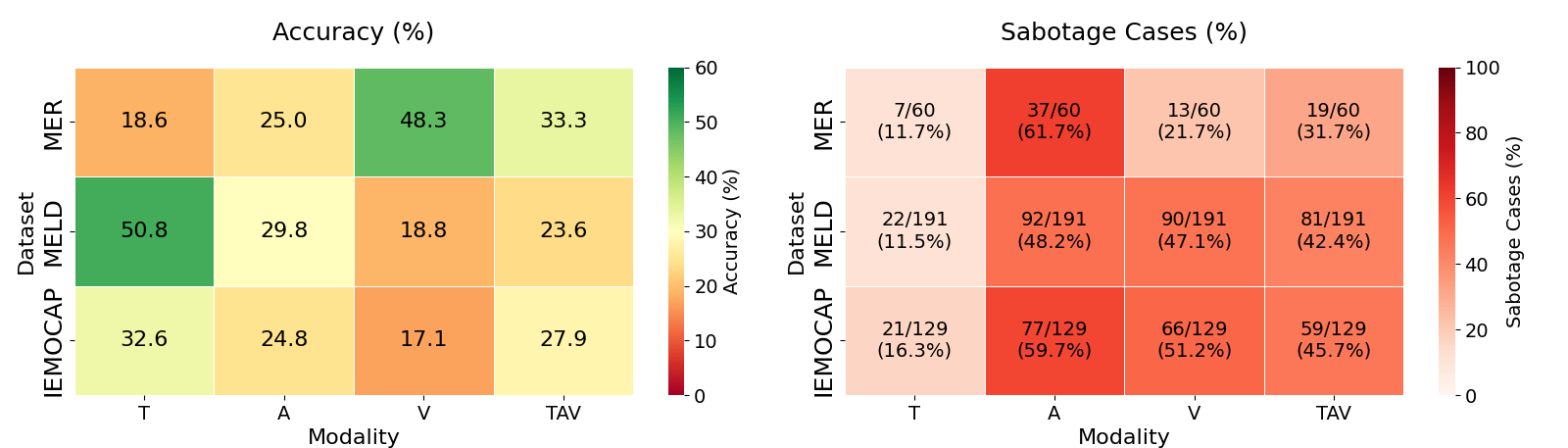}
     \caption{\textit{Left heatmap}: unimodal accuracy for Text (T), Audio (A), Vision (V), and joint view (TAV), highlighting differences across datasets. 
    \textit{Right heatmap}: proportion of cases where a modality \emph{sabotages} the fused decision 
    (high-confidence error flipping Top-1 at threshold 70), where each values show \#cases/total (rate\%). 
    }
    \label{fig:sabotage_bars}
\end{figure*}

\vspace{-2pt}
We operationalize \emph{modality sabotage} as a measurable, instance‑level diagnostic for
\emph{high‑confidence, misleading} unimodal outputs that dominate the fusion and derail the final
decision. This test makes the notion of “pulling the decision away” explicit, yields a
\emph{countable event} per example, and supports auditing by answering \emph{who contributed} or
\emph{who hurt} each prediction.
Figure~\ref{fig:sabotage_bars} visualizes unimodal accuracy and sabotage rates per modality for
GPT‑5‑nano under confidence‑weighted fusion.

\paragraph{Diagnostic signals revealed.}
Beyond aggregate rates, the sabotage test surfaces actionable signals at the instance level as reported in Figure~\ref{fig:sabotage_bars}: 
(i)Across the columns, we observe a per-modality \emph{calibration gap} (self-reported confidence vs.\ empirical accuracy), and 
(ii)Across the rows, we report the \emph{dataset/backbone reliability profiles} ranking modalities by accuracy and sabotage. 
\textit{Comparing the columns} in Figure~\ref{fig:sabotage_bars} (Left), patterns are consistent: audio is the primary saboteur and text most contributed. 
This provides a basis for identifying which components of a model pipeline may require refinement. 
\textit{Across rows}, we can evaluate which modalities are less reliable within each dataset. This is consistent with each dataset characteristics: \textbf{MER} suffers from noisy ASR/translation but benefits from rich video cues; \textbf{MELD}'s sitcom-style video with exaggerated cues or multiple actors can mislead vision; \textbf{IEMOCAP} features seated dyads, where acted expressions and experimental scenes limit visual reliability.

\vspace{-5pt}

\section{Conclusion}
We presented a lightweight, model-agnostic diagnostic framework that makes multimodal fusion decisions interpretable at the instance level. Central to our analysis is \emph{modality sabotage}, a failure mode in which a high-confidence unimodal error misleads the fused prediction. Our results demonstrate that this framework can expose systematic reliability patterns and recoverable uncertainty across datasets and backbones. Beyond emotion recognition, the proposed approach offers a general scaffold for auditing multimodal reasoning systems and guiding future work on calibration, conflict resolution, and interpretable fusion.

\bibliographystyle{unsrtnat}
\bibliography{main}






\end{document}